\documentclass[conference]{IEEEtran}
\IEEEoverridecommandlockouts
\usepackage{cite}
\usepackage{amsmath,amssymb,amsfonts}
\usepackage{algorithmic}
\usepackage{graphicx}
\usepackage{textcomp}
\usepackage{xcolor}

\usepackage{subcaption}
\usepackage[normalem]{ulem}

\usepackage{xcolor}

\usepackage[utf8]{inputenc}
\usepackage{fourier} 
\usepackage{array}
\usepackage{makecell}

\usepackage{caption}

\def\BibTeX{{\rm B\kern-.05em{\sc i\kern-.025em b}\kern-.08em
    T\kern-.1667em\lower.7ex\hbox{E}\kern-.125emX}}

\begin{document}




\title{Autoencoder Based Face Verification System}

\author{\IEEEauthorblockN{Enoch Solomon}
\IEEEauthorblockA{\textit{Department of Computer Science} \\
Virginia State University \\
Richmond, Virginia \\
esolomon@vsu.edu
}

\and
\IEEEauthorblockN{Abraham Woubie}
\IEEEauthorblockA{\textit{Silo AI} \\
Helsinki, Finland\\
abraham.zewoudie@silo.ai}

\and

\IEEEauthorblockN{Eyael Solomon Emiru}
\IEEEauthorblockA{\textit{Department of Information} \\
Engineering and Computer Science \\
University of Trento, Italy \\
eyael.emiru@studenti.unitn.it}
}

\maketitle

\begin{abstract}

The primary objective of this work is to present an alternative approach aimed at reducing the dependency on labeled data. Our proposed method involves utilizing autoencoder pre-training within a face image recognition task with two step processes. Initially, an autoencoder is trained in an unsupervised manner using a substantial amount of unlabeled training dataset. Subsequently, a  deep learning model is trained with initialized parameters from the pre-trained autoencoder. This deep learning training process is conducted in a supervised manner, employing relatively limited labeled training dataset. During evaluation phase, face image embeddings is generated as the output of deep neural network layer. Our training is executed on the CelebA dataset, while evaluation is performed using benchmark face recognition datasets such as Labeled Faces in the Wild (LFW) and YouTube Faces (YTF). Experimental results demonstrate that by initializing the deep neural network with pre-trained autoencoder parameters  achieve comparable results to state-of-the-art methods.
\end{abstract}


\section{Introduction} 

Face verification serves as a means to gain entry to an application, system, or service. Its primary objective is to assess a provided facial image in comparison to another face and confirm if they correspond. Essentially, when presented with two facial images, the face verification algorithm determines whether they depict the same individual. This method, distinct from traditional verification techniques like passwords or fingerprints, relies on dynamic patterns within biometric face verification, making it exceptionally secure and efficient. Beyond access control, face recognition finds applications in forensics and transaction authentication.

\begin{figure*}[h!]
	\centering
		\includegraphics[scale=0.8]{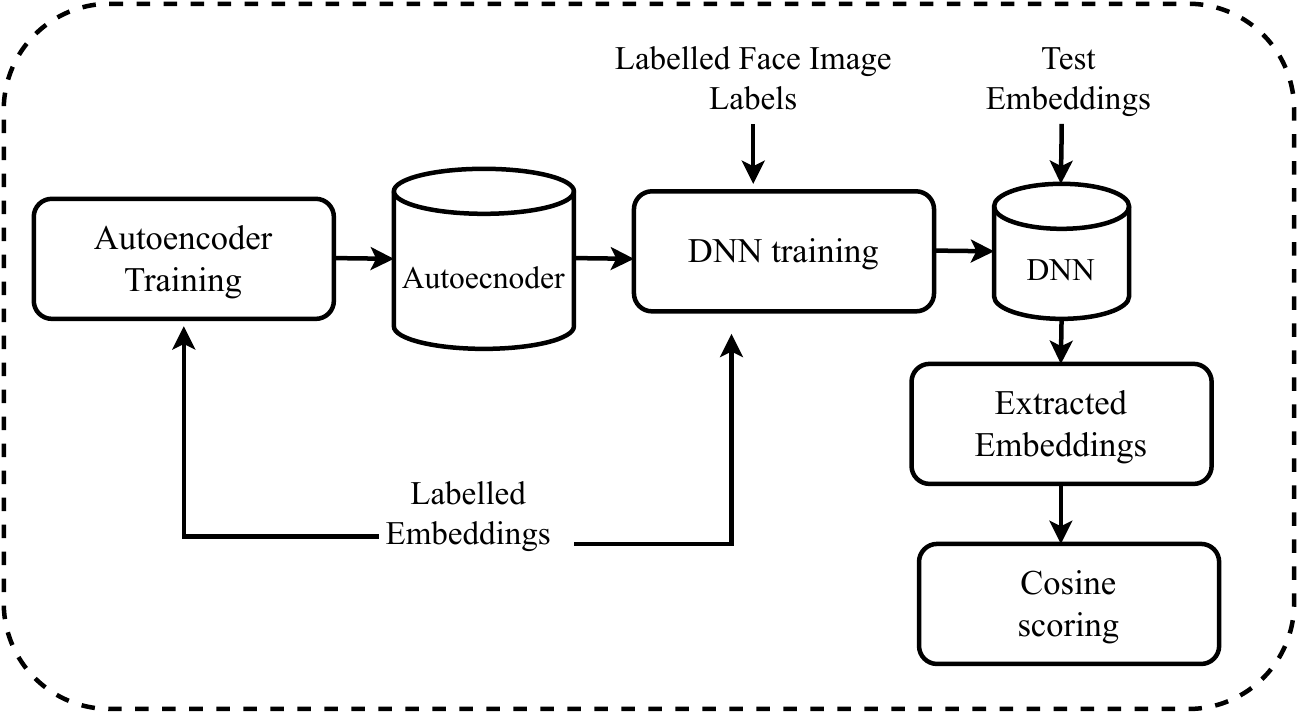}
	\caption{The proposed Face embedding extraction from autoencoder pre-training.}
	\label{fig:proposed}
\end{figure*}

The utilization of deep neural networks in face verification, as demonstrated in studies such as \cite{schroff2015facenet,taigman2014deepface,taigman2015web,deng2017marginal,deng2019arcface,kim2020groupface,huang2020curricularface,deng2021variational,solomon2023deep,solomon2023unsupervised,woubie2023image,https://doi.org/10.25772/re06-av14}, has exhibited significant advancements in face verification systems. These deep neural networks have also proven successful in various speech applications and face anti-spoofing scenarios \cite{woubie2021federated,woubie2021voice,woubie2021federatedondevice,woubie2021use,solomon2023fass,solomon2023hdlhc}.

One notable example is the Facenet system \cite{schroff2015facenet}, developed by Google, which employed a Siamese network trained on a labeled dataset comprising 200 million faces. It achieved a remarkable accuracy of over 98\% on the LFW dataset \cite{huang2008labeled} and over 95\% on YTF \cite{wolf2011face}, both benchmark face verification datasets. DeepFace \cite{taigman2014deepface}, developed by Meta, utilized 3D face modeling and a nine-layer network with approximately 120 million parameters. Trained on 4.4 million labeled face images, DeepFace achieved an accuracy of over 97\% on the LFW dataset. Further enhancements were made in \cite{taigman2015web} by incorporating over 500 million face images for training, resulting in an improved performance of over 98

In another approach, a face verification system proposed in \cite{deng2017marginal} utilized marginal loss and was trained on a 4 million labeled dataset, achieving an accuracy of over 99\% on LFW and over 95\% on YTF. ArcFace \cite{deng2019arcface} introduced an additive angular margin loss, surpassing 99\% accuracy on LFW and over 98\% on YTF. GroupFace \cite{kim2020groupface}, leveraging multiple group-aware representations, achieved over 99\% and 97\% on the LFW and YTF datasets, respectively, although both ArcFace and GroupFace required a labeled training dataset of 5.8 million samples.

CurricularFace \cite{huang2020curricularface} implemented adaptive curriculum learning loss, surpassing 99\% accuracy on the LFW dataset. MDCNN composed of two advanced deep learning neural network models, achieved over 99\% and 94\% on the LFW and YTF datasets, respectively, using a 1 million labeled training dataset. 

Semi-supervised and unsupervised learning techniques in the context of deep neural networks typically employ one of two primary approaches. The first one is consistency regularization-based method that incorporates a regularization term into the objective function, aiming to enforce consistency during training, especially when dealing with a substantial amount of unlabeled data. This regularization constrains the model's predictions to remain stable or invariant when faced with input noise or variations. They  introduced an Unsupervised Domain Adaptation approach, which utilizes advanced data augmentation techniques like rand-augment and back-translation to enhance model robustness. The second one is proxy label-based methods. In the second approach, pseudo-labels, often referred to as proxy labels, are initially assigned to the unlabeled data. Subsequently, training occurs based on these proxy labels in conjunction with the genuine ground-truth labels for both labeled and unlabeled data. This method effectively leverages these proxy labels to guide the model's learning process. These two strategies represent key avenues for semi-supervised learning with deep neural networks, each offering distinct advantages and trade-offs in various applications. The research by \cite{fang2023rethinking} introduced a silhouette coefficient-based contrast clustering algorithm. This innovative approach assesses the degree of separation between clusters by scrutinizing both intra-cluster and inter-cluster distances, thereby enabling precise detection of out-of-distribution data. The work in \cite{liu2023discriminative} proposed discriminative sparse least square regression for semi-supervised learning. The authors in ~\cite{liu2021unsupervised,boutros2023unsupervised} explore face synthesis and pose-invariant face recognition using generative adversarial networks (GANs).

By relying on unsupervised networks, researchers can leverage the intrinsic structure and patterns within the data to create tailored representations for each person. This approach has the potential to improve recognition accuracy, especially in scenarios where labeled data for each individual may be scarce or impractical to obtain.

In summary, unsupervised deep learning methods, such as RBMs, DBNs, and autoencoders, offer a promising avenue for face image recognition that doesn't necessitate labeled data for all individuals. Instead, they enable the training of separate models for each target person, thereby harnessing the power of unsupervised learning to enhance recognition performance.
 
Many of the supervised methods mentioned above necessitate extensive labeled training data, a resource often lacking in numerous domains. Moreover, the challenge of gathering sufficient labels for real-world recognition applications further complicates matters. Consequently, the performance of these methods experiences a significant decline in such scenarios.

In this work, our primary aim is to alleviate the demand for labeled data within face image verification systems. To accomplish this, we introduce an innovative approach that incorporates an autoencoder pre-training model to extract DNN-based face image embeddings for face verification tasks. Our approach centers on bypassing the need for a large quantity of labeled data. Instead, we kickstart the autoencoder training with a substantial volume of unlabeled data. Following this, deep learning neural network is trained using a comparatively modest amount of labeled training dataset. The key to our strategy lies in initializing the deep neural network training process with parameters acquired from the previous steps of pre-trained autoencoder. This method enables the creation of a hybrid autoencoder deep neural network classifier, combining the strengths of both components.

Following the training phase, we extract face image embeddings from the input image vector, during evaluation. Our overarching objective is to enhance system performance while minimizing the reliance on labeled face image training dataset. The results of our experiments highlight two significant advancements. Firstly, the face image embeddings generated using our proposed method, coupled with cosine scoring, demonstrate accuracy levels comparable to those achieved by state-of-the-art supervised methods. Secondly, we observe that the proposed method training process converges at a faster rate when compared to the classical deep neural network training without the benefit of autoencoder pre-training.

The paper unfolds in the following structure: Section 2 elucidates the proposed method, delving into its intricacies. Moving forward, Section 3 delineates the experimental setup and details the data employed. Section 4 delves into a comprehensive discussion of the results obtained. Bringing the discourse to a close, Section 5 encapsulates the findings and draws conclusions.

\section{Proposed Method}

In this work, we introduce a novel framework for face image verification by employing autoencoder to generate an embedding vector of facial images. In contrast to conventional DNN classifiers, we circumvent the need for extensive labeled data. Fig. \ref{fig:proposed}  illustrates the architecture of our proposed process for extracting face image embeddings. Initially, we train an autoencoder, leveraging a substantial corpus of unlabeled facial images. 
In this training phase, we adhere to a conventional approach, aiming to minimize the Mean Square Error (MSE) loss between the input and the reconstructed facial image vector. We employ the Stochastic Gradient Descent (SGD) optimizer to facilitate this optimization process.

After autoencoder training, deep neural network classifier is trained in a supervised manner with little labeled training dataset to acquire knowledge about individual persons face. This process involves expanding upon the autoencoder's architecture by adding a fully connected layer and a classification layer after its final layer.  The proposed deep neural network is initialized with parameters from the previous steps of pre-trained autoencoder. The autoencoder pre-training helps in supervised learning by providing an initial representation of data and it also enables the network to converge relatively faster with pre-training compared to training from scratch.

There are two discernible scenarios for leveraging the pretrained autoencoder for deep neural network initialization. One approach involves integrating the fully connected and classification layers directly following the encoder section. The encoder part compresses the data into a lower-dimension, retaining enough information to reconstruct original data. However, our experimental findings caution against this method. In the proposed deep neural network training method, the network benefits from additional information gleaned from the face image labels at the output. The encoder's limited dimensional space proves inadequate for effectively assimilating knowledge from the higher-dimensional classification layer.

Another approach involves utilizing the entire autoencoder by integrating the fully connected and classification layers at the end of the autoencoder architecture. Our preference is to restore the input data to its original dimensional space before training the hybrid autoencoder deep neural network classifier. This two-step process yields several advantages. Firstly, it provides us dimensionality reduction. Initially, we utilize the encoder to compress the input data into a lower-dimension, effectively removing extraneous information. Secondly, it enables  information expansion. Subsequently, we expand the data back to its original dimensional space. This expansion facilitates the acquisition of additional information from the face image labels, enhancing the network's ability to learn discriminative features.

This approach allows us to strike a balance between dimensionality reduction and information retention, ultimately leading to more effective learning during the hybrid autoencoder deep neural network training process.

\begin{figure}[]
	\centering
		\includegraphics[scale=0.4]{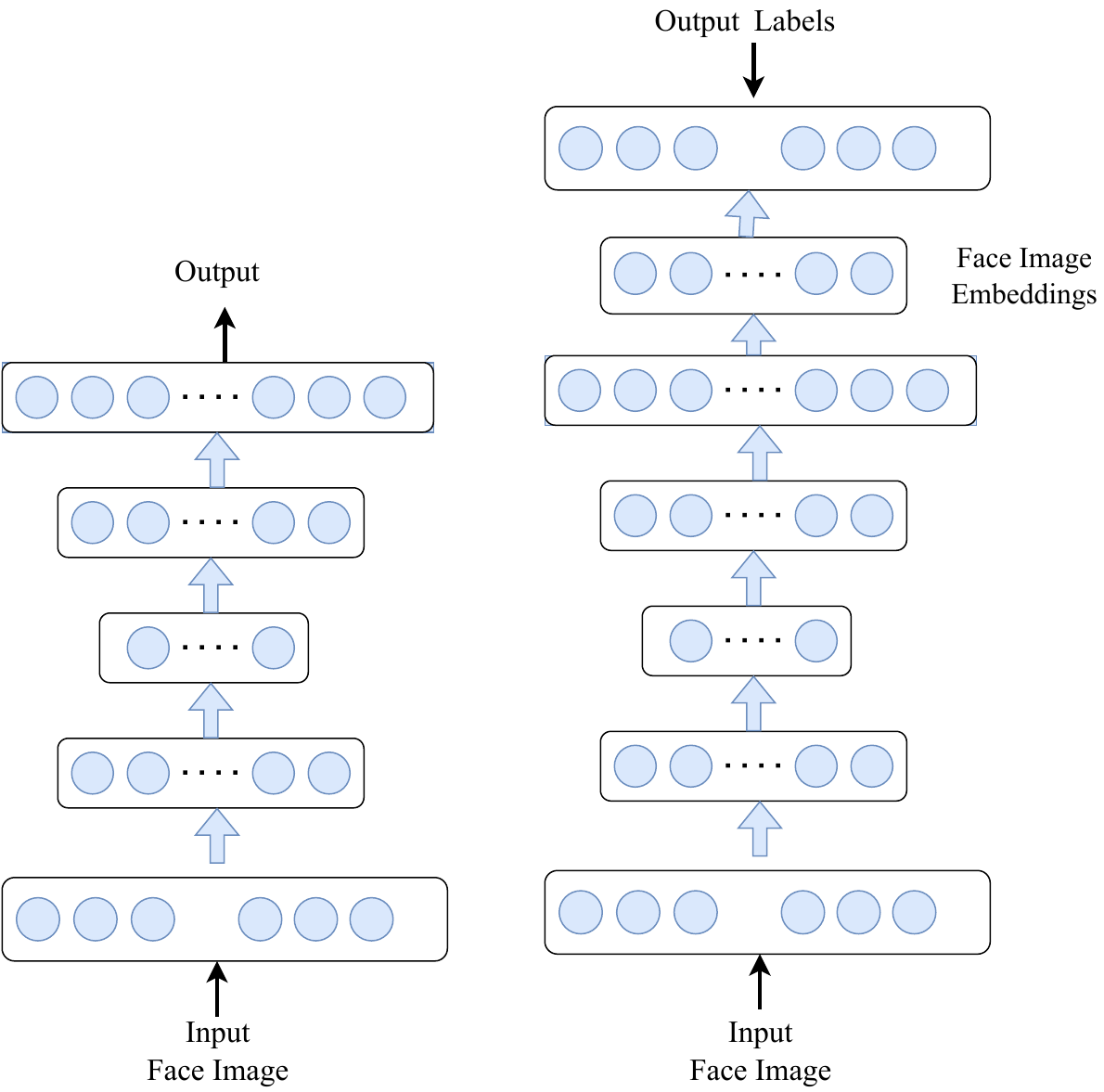}
	\caption{The left side is an autoencoder pre-training whereas the right side is with a DNN training.}
	\label{fig:arch2}
\end{figure}

Fig. \ref{fig:arch2} illustrates the architectural designs of both the autoencoder and the DNN. The autoencoder features a symmetrical structure, with an encoder part and a decoder part, resembling the conventional autoencoder configuration. Meanwhile, the deep neural network maintains a similar structural symmetry.

Lastly, the output obtained from the last layer serves as the sought-after face image embeddings, which have demonstrated their capacity to preserve individual-specific face information effectively. During the evaluation phase, test face images embedding vectors are extracted. Subsequently, these embeddings are employed in the experimentation trials, employing the cosine scoring technique for evaluation purposes.

\section{Experimental Setup and Dataset}

\subsection{Experimental Setup}

For model training, we employed the Keras deep learning library  \cite{chollet2015keras}. The architecture of the autoencoder employed in this research comprises three hidden layers. Symmetry is maintained between the encoder and decoder sections, with both the first and third hidden layers containing an identical number of neurons, as depicted in Fig. 2. Each layer in the encoder input and decoder output comprises 112 by 112 neurons. Moving to the second layer of both the encoder and decoder, there are 800 neurons. The encoder's output layer, crucial in determining the size of the resulting embedding vector, is composed of 300 neurons. Transitioning to the DNN, the dimension of the face image embedding layer remains fixed at 400, while the classification layer encompasses 1000 neurons.

The autoencoder is training  for 500 epochs or until the error ceased to decrease. The final layer of autoencoder is using linear activation function whereas the other layers are using ReLU activation function. The learning rate is set at 0.003, with a decay rate of 0.00002, and a logarithmically decaying learning rate spanning from $10^{-3}$ to $10^{-9}$. The batch size was configured at 100 for this phase.

The supervised training of the deep neural network was run for 300 epochs or until the error no longer exhibited a decreasing trend. We utilized the SGD with momentum optimizer with learning rate of 0.002. Sigmoid activation was applied to all layers within the deep learning architecture.

\subsection{Datasets}
The CelebA dataset \cite{liu2015deep} boasts a collection of over 200,000 images featuring over 10,000 celebrities, encompassing diverse pose variations and background clutter. To facilitate model training and evaluation, the dataset is partitioned into training, validation, and test sets. In our study, we harnessed the training and test segments of the CelebA dataset as an unlabeled dataset for training the autoencoder. The supervised training, on the other hand, was conducted using the validation portion of the CelebA dataset.

The Labeled Faces in the Wild dataset, abbreviated as \textbf{(LFW)} \cite{huang2008labeled}, encompasses over 13,000 images featuring over 5,000 individuals. In the testing phase, the database undergoes a random split into 10 subsets. Subsequently, within each subset, 300 matched pairs and 300 mismatched pairs are randomly selected. For testing purposes, a total of 3000 matched pairs and 3000 mismatched pairs \cite{huang2008labeled} are utilized.

The YouTube Faces dataset, denoted as \textbf{(YTF)} \cite{wolf2011face}, comprises more than 3,000 face videos featuring over 1,000 individuals sourced from YouTube. On average, each person is represented by two videos. The video clips vary in duration, with the shortest consisting of 46 frames and the longest spanning over 6,000 frames. The average length of a video clip in the dataset is about 180 frames. For testing purposes, 5,000 video pairs are randomly selected and prepared. Half of these pairs depict videos of the same person, while the other half feature videos of different individuals. In essence, for testing, a total of 5,000 pairs of static images are employed, with 2,500 of them depicting the same person and 2,500 showcasing different individuals \cite{wolf2011face}.

\begin{figure*}[h!]
	\centering
		\includegraphics[width=17.5 cm]{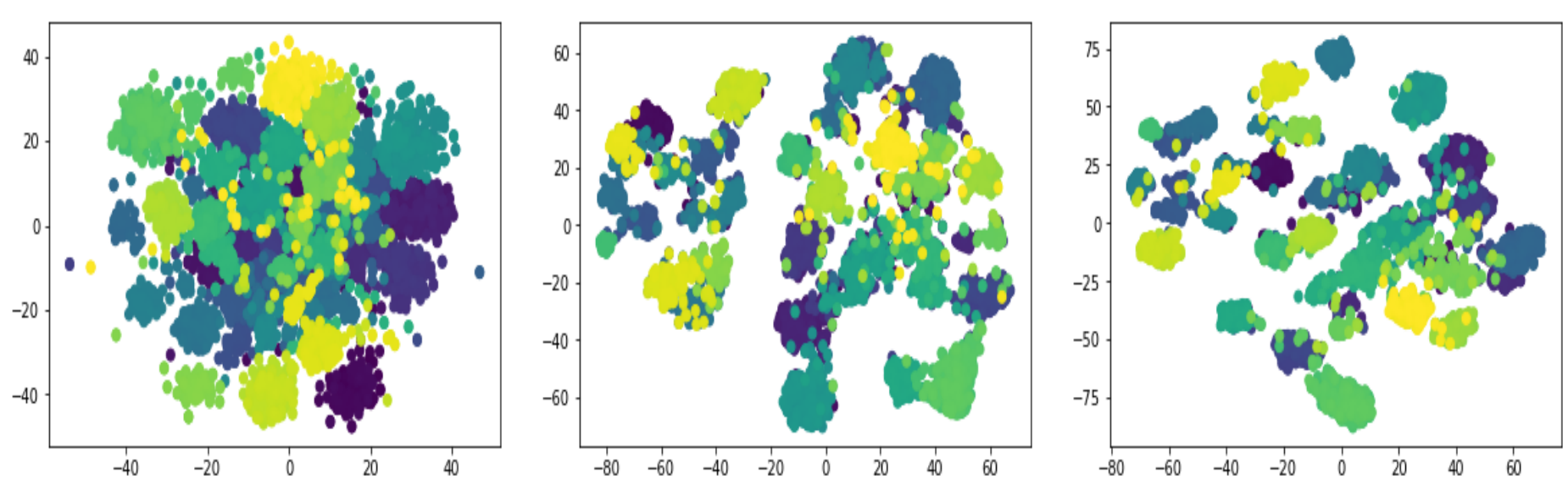}
	\caption{The most left one is the raw face images, the middle one is the conventional DNN face image embeddings and the right most one is the proposed face image embeddings.}
	\label{fig:attack}
\end{figure*}

\section{Results}
\subsection{Results on LFW}

\begin{table}[]
\caption{Comparison of the proposed method, which utilizes a Siamese network with some state-of-the-art methods on the LFW testing dataset.}
\label{table:experimental_result_LFW}
\begin{tabular}{lllll}
\hline
\thead{Model} & \thead{Training \\ size} &  \thead{Labeled/ \\ Unlabeled} &  \thead{Testing \\ size} & \thead{Accuracy(\%)} \\ \hline

ArcFace               &  5.8M                           & Labeled   & 6K                                              & 99.82 \cite{deng2019arcface}                        \\ \hline

GroupFace  & 5.8M & labeled & 6K   & 99.85 \cite{kim2020groupface}                                    \\ \hline

DeepFace-ensemble               & 4.4M                            & Labeled   & 6K                                             & 97.35 \cite{taigman2014deepface}                       \\ \hline

 Stream Loss               &  1.5M                           & Labeled     & 6K                                          & 98.97 \cite{rashedi2019stream}                        \\ \hline

 F$^2$C               &  5.8M                          & Labeled   & 6K                                             & 99.83 \cite{wang2022efficient}                       \\
 \hline

USynthFace               &  1M                          & Unlabeled    & -                                            & 92.23 \cite{boutros2023unsupervised}                        \\

 \hline
 UFace               &  200K                          & Unlabeled    & 6K                                            & 99.40 \cite{solomon2022uface}                        \\

 \hline
 
\textbf{Proposed}      & \textbf{200K}                             & \textbf{Unlabeled}    & \textbf{6K}                                            & \textbf{99.60}              \\ \hline

\end{tabular}
\end{table}

Table \ref{table:experimental_result_LFW} presents a comparison of our proposed method with state-of-the-art approaches. It's important to highlight that our comparison encompasses results from methods that employ both supervised and unsupervised training. Notably, our proposed training method doesn't require an extensive amount of labeled data.

Indeed, while methods like ArcFace and GroupFace may exhibit slightly higher accuracy, it's crucial to emphasize that our proposed method achieves commendable performance despite being trained on a significantly smaller dataset of around 200,000 images. In contrast, many state-of-the-art methods rely on training with millions of images. This underscores the efficiency and efficacy of our approach, showcasing competitive results with a more modest dataset.

The experiments revealed that autoencoder pre-training successfully captured person-specific facial features from a large pool of unlabeled face images. This acquired knowledge was effectively utilized by the DNN classifier, benefiting from the initialization with autoencoder weights and biases. Consequently, a notable enhancement in the convergence of DNN training was observed.

Notably, slight improvement was observed when comparing the performance of deep neural network trained with only the encoder versus those initialized with the entire autoencoder. However, the optimal choice was found to be the full autoencoder initialization, which significantly expedited the training convergence. Beyond some epochs, the validation loss exhibited a considerably lower absolute value compared to conventional training.

Following the extraction of desired face image embeddings using our proposed approach, testing were done using the cosine scoring technique. Table I provides a comparative analysis, highlighting the performance of our proposed face image embeddings against state-of-the-art face verification methods. It's noteworthy that utilizing only the encoder for DNN training proved to be suboptimal in our experiments. The complete autoencoder initialization, leveraging information from face image labels, demonstrated a clear advantage over the encoder-only approach. The results in Table I underscore that our proposed face image embedding is comparable with the performance of most state-of-the-art face verification methods.

\subsection{Results on YTF}

\begin{table}[]
\caption{Comparison of the proposed method, which utilizes a Siamese network with some state-of-the-art methods on the YTF testing dataset.}
\label{table:experimental_results_YTF}
\centering
\begin{tabular}{lllll}
\hline
\thead{Model} & \thead{Training \\ size} & \thead{Labeled/ \\ Unlabeled}  & \thead{Testing \\ size} & \thead{Testing \\ accuracy(\%)} \\ \hline

ArcFace  & 5.8M & labeled & 5K & 98.02 \cite{deng2019arcface}                                    \\ \hline
GroupFace  & 5.8M & labeled & 5K  & 97.80 \cite{kim2020groupface}                                                                       \\ \hline
DeepFace-single & 4.4M & labeled & 5K & 91.40 \cite{taigman2014deepface}                        \\ \hline

Stream Loss  & 1.5M & Labeled & 5K & 96.40 \cite{rashedi2019stream}                                    \\ \hline

F$^2$C  & 1M & Labeled & 5K & 97.76 \cite{wang2022efficient}                                   \\ \hline

UFace  & 200K & Unlabeled & 5K & 96.04 \cite{solomon2022uface}                                    \\ \hline

 \textbf{Proposed } & \textbf{200K} & \textbf{Unlabeled} & \textbf{5K} & \textbf{96.82}                    \\ \hline
\end{tabular}
\end{table}

We conducted a comparative analysis of our proposed method against state-of-the-art supervised and unsupervised approaches on the YTF dataset. In Table \ref{table:experimental_results_YTF}, Deng et al. \cite{deng2017marginal} utilized marginal loss with a labeled 4 million training dataset, achieving results comparable to Facenet \cite{schroff2015facenet}, which employed a larger 200 million labeled training dataset and achieved over 95\% accuracy. Notably, these methods rely on extensive labeled datasets for training.

Contrastingly, our proposed method achieves commendable results using significantly less unlabeled training data and a limited amount of labeled data, surpassing 96\% accuracy. When compared with both state-of-the-art supervised and unsupervised methods on YTF, our method's accuracy (96.04\%) slightly outperforms certain supervised methods, surpasses unsupervised methods, and closely approaches the performance of leading methods such as ArcFace and GroupFace.

In Fig. \ref{fig:attack}, we present t-Distributed Stochastic Neighbor Embedding plots illustrating three different vector representations of face images: raw face images, conventional DNN face image embeddings, and our proposed face image embeddings. T-SNE is employed as a dimensionality reduction technique to visually represent higher-dimensional vectors \cite{van2008visualizing}. To gauge the discriminative capability of our proposed face image embeddings, we compare the t-SNE plots with the other two approaches. The plots are derived from the test partition of the LFW dataset, with all vectors reduced to 2 dimensions for visualization.

The figure clearly demonstrates that our proposed face image embeddings exhibit good discriminative power. The generated clusters are predominantly distinct, emphasizing the efficacy of our approach. In contrast, the t-SNE plots for conventional DNN face image embeddings show some clusters overlapping with others, indicating a lower level of discrimination. 

\section{Conclusion}

This work introduces an innovative approach to face recognition, employing autoencoder pretraining for generating Deep Neural Network (DNN) face image embeddings in the context of face verification tasks. The demanding requirement for abundant labeled data has posed a significant challenge for deep learning methodologies in this domain.  In real-world scenarios, access to large quantities of labeled data is often limited. Therefore, we turn to the utilization of unlabeled dataset to mitigate the impact of this data scarcity.

The proposed method starts by pre-training an autoencoder on an extensive unlabeled training dataset, enabling it to learn facial features that are independent of specific individuals. Subsequently, we train DNN classifier using a relatively modest amount of labeled training dataset, initializing it with the weights and bias derived from the previous steps which is pre-trained autoencoder. During experimentation, we extract face image embeddings from input images.

We conducted evaluations using well-established face image verification benchmark datasets like LFW and YTF. The results illustrate that by employing autoencoder pretraining for DNN, we achieve performance levels comparable to many state-of-the-art supervised face verification methods. Additionally, we observed that the DNN training process converges more swiftly when compared to the classical DNN training.


\begin{thebibliography}{00}
\bibitem{schroff2015facenet}Schroff, F., Kalenichenko, D. \& Philbin, J. Facenet: A unified embedding for face recognition and clustering. {\em Proceedings Of The IEEE Conference On Computer Vision And Pattern Recognition}. pp. 815-823 (2015)
\bibitem{taigman2014deepface}Taigman, Y.; Yang, M.; Ranzato, M.; Wolf, L. Deepface: Closing the gap to human-level performance in face verification. {{In Proceedings of the} IEEE Conference On Computer Vision And Pattern Recognition}, {Columbus, OH, USA, 23-28 June} 2014; pp. 1701--1708.
\bibitem{taigman2015web}Taigman, Y., Yang, M., Ranzato, M. \& Wolf, L. Web-scale training for face identification. {\em Proceedings Of The IEEE Conference On Computer Vision And Pattern Recognition}. pp. 2746-2754 (2015)

\bibitem{huang2020curricularface}Huang, Y., Wang, Y., Tai, Y., Liu, X., Shen, P., Li, S., Li, J. \& Huang, F. Curricularface: adaptive curriculum learning loss for deep face recognition. {\em Proceedings Of The IEEE/CVF Conference On Computer Vision And Pattern Recognition}. pp. 5901-5910 (2020)
\bibitem{deng2021variational}Deng, J., Guo, J., Yang, J., Lattas, A. \& Zafeiriou, S. Variational prototype learning for deep face recognition. {\em Proceedings Of The IEEE/CVF Conference On Computer Vision And Pattern Recognition}. pp. 11906-11915 (2021)
\bibitem{https://doi.org/10.25772/re06-av14}Solomon, E. Face Anti-Spoofing and Deep Learning Based Unsupervised Image Recognition Systems. {\em VCU Theses And Dissertations}. (2023), https://scholarscompass.vcu.edu/etd/7482

\bibitem{solomon2023deep}Solomon, E., Woubie, A. \& Emiru, E. Deep Learning Based Face Recognition Method using Siamese Network. {\em ArXiv Preprint ArXiv:2312.14001}. (2023)
\bibitem{solomon2023unsupervised}Solomon, E., Woubie, A. \& Emiru, E. Unsupervised Deep Learning Image Verification Method. {\em ArXiv Preprint ArXiv:2312.14395}. (2023)

\bibitem{woubie2023image}Woubie, A., Solomon, E. \& Emiru, E. Image Clustering using Restricted Boltzman Machine. {\em ArXiv Preprint ArXiv:2312.13845}. (2023)


\bibitem{deng2017marginal}Deng, J., Zhou, Y. \& Zafeiriou, S. Marginal loss for deep face recognition. {\em Proceedings Of The IEEE Conference On Computer Vision And Pattern Recognition Workshops}. pp. 60-68 (2017)
\bibitem{deng2019arcface}Deng, J., Guo, J., Xue, N. \& Zafeiriou, S. Arcface: Additive angular margin loss for deep face recognition. {\em Proceedings Of The IEEE/CVF Conference On Computer Vision And Pattern Recognition}. pp. 4690-4699 (2019)

\bibitem{kim2020groupface}Kim, Y., Park, W., Roh, M. \& Shin, J. Groupface: Learning latent groups and constructing group-based representations for face recognition. {\em Proceedings Of The IEEE/CVF Conference On Computer Vision And Pattern Recognition}. pp. 5621-5630 (2020)


\bibitem{woubie2021federated} Abraham Woubie and Tom Bäckström: Federated learning for privacy-preserving speaker recognition, { In IEEE Access,} 2021; pp. 149477--149485.
\bibitem{woubie2021voice} Abraham Woubie and Tom Bäckström: Voice-quality Features for Deep Neural Network Based Speaker Verification Systems, {In 29th European Signal Processing Conference (EUSIPCO),} 2021; pp. 176--180.
\bibitem{woubie2021federatedondevice} Abraham Woubie and Tom Bäckström: Federated Learning for Privacy Preserving On-Device Speaker Recognition, {In ISCA Symposium on Security and Privacy in Speech Communication,} 2021; pp. 1--5.
\bibitem{woubie2021use} Abraham Woubie, Pablo Zarazaga and Tom Bäckström: The Use of Audio Fingerprints for Authentication of Speakers on Speech Operated Interfaces, {In ISCA Symposium on Security and Privacy in Speech Communication,} 2021; pp. 6--9.
\bibitem{solomon2023fass}Solomon, E. \& Cios, K. FASS: Face Anti-Spoofing System Using Image Quality Features and Deep Learning. {\em Electronics}. \textbf{12}, 2199 (2023)
\bibitem{solomon2023hdlhc}Solomon, E. \& Cios, K. HDLHC: Hybrid Face Anti-Spoofing Method Concatenating Deep Learning and Hand-Crafted Features. {\em 2023 IEEE 6th International Conference On Electronic Information And Communication Technology (ICEICT)}. pp. 470-474 (2023)
\bibitem{huang2008labeled}Huang, G., Mattar, M., Berg, T. \& Learned-Miller, E. Labeled faces in the wild: A database forstudying face recognition in unconstrained environments.  (2008)
\bibitem{wolf2011face}Wolf, L., Hassner, T. \& Maoz, I. Face recognition in unconstrained videos with matched background similarity. {\em CVPR 2011}. pp. 529-534 (2011)

\bibitem{fang2023rethinking}Fang, B., Li, X., Han, G. \& He, J. Rethinking pseudo-labeling for semi-supervised facial expression recognition with contrastive self-supervised learning. {\em IEEE Access}. (2023)

\bibitem{liu2023discriminative}Liu, Z., Lai, Z., Ou, W., Zhang, K. \& Huo, H. Discriminative sparse least square regression for semi-supervised learning. {\em Information Sciences}. \textbf{636} pp. 118903 (2023)

\bibitem{liu2021unsupervised}Liu, Y.; Chen, J. Unsupervised face frontalization for pose-invariant face recognition. {\em Image Vis. Comput.} \textbf{2021}, \emph{106}, 104093.
\bibitem{boutros2023unsupervised}Boutros, F., Klemt, M., Fang, M., Kuijper, A. \& Damer, N. Unsupervised face recognition using unlabeled synthetic data. {\em 2023 IEEE 17th International Conference On Automatic Face And Gesture Recognition (FG)}. pp. 1-8 (2023)

\bibitem{chollet2015keras}Chollet, F. \& Others Keras. https:// keras. io.  (2015)
\bibitem{liu2015deep}Liu, Z., Luo, P., Wang, X. \& Tang, X. Deep learning face attributes in the wild. {\em Proceedings Of The IEEE International Conference On Computer Vision}. pp. 3730-3738 (2015)



\bibitem{rashedi2019stream}Rashedi, E., Barati, E., Nokleby, M. \& Chen, X. “Stream loss”: ConvNet learning for face verification using unlabeled videos in the wild. {\em Neurocomputing}. \textbf{329} pp. 311-319 (2019)

\bibitem{wang2022efficient}Wang, K.; Wang, S.; Zhang, P.; Zhou, Z.; Zhu, Z.; Wang, X.; Peng, X.; Sun, B.; Li, H.; You, Y. An efficient training approach for very large scale face recognition. {{In Proceedings of the} IEEE/CVF Conference on Computer Vision and Pattern Recognition}, {New Orleans, Louisiana, USA, 21-24 June} 2022; pp.~4083--4092.

\bibitem{solomon2022uface}Solomon, E., Woubie, A. \& Cios, K. UFace: An Unsupervised Deep Learning Face Verification System. {\em Electronics}. \textbf{11}, 3909 (2022)

\bibitem{van2008visualizing}Van, der. Maaten.; Laurens; Hinton, Geoffrey. Visualizing data using t-SNE. {Journal of machine learning research} AAAI Conference on Artificial Intelligence, 2008; Volume {9}.

\end{thebibliography}
\end{document}